\documentclass[12pt]{article}
\usepackage{cite}
\usepackage{amsmath,amssymb,amsfonts}
\usepackage{algorithmic}
\usepackage{graphicx}
\usepackage{algorithm,algorithmic}
\usepackage{float}
\usepackage{dblfloatfix}
\usepackage{hyperref}
\hypersetup{hidelinks=true}
\usepackage{textcomp}
\usepackage{array}
\usepackage{xcolor}
\usepackage{hyperref}
\usepackage{textcomp}
\usepackage{upquote}
\usepackage{subfig}
\usepackage{multirow}

\newcolumntype{P}[1]{>{\centering\arraybackslash}p{#1}}

\usepackage{color,array}

\usepackage{graphicx}

\begin{document}

\title{Exploring Topic Trends in COVID-19 Research Literature using Non-Negative Matrix Factorization}

\author{Divya Patel\thanks{Divya Patel, Vansh Parikh, Om Patel and Bhaskar Chaudhury are with Group in Computational Science and HPC, Dhirubhai Ambani Institute of Information and Communication Technology, Gandhinagar, India (e-mails: divyapatel0273@gmail.com, vanshparikh20112002@gmail.com, o.v.patel2705@gmail.com, bhaskar\_chaudhury@daiict.ac.in).} , Vansh Parikh$^{*}$, Om Patel$^{*}$,\\ Agam Shah\thanks{Agam Shah is with the School of Computational Science \& Engineering, College of Computing, Georgia Institute of Technology, Atlanta, GA, USA (e-mail: ashah482@gatech.edu).} , Bhaskar Chaudhury$^{*}$ 
}

\markboth{Journal of IEEE Transactions on Artificial Intelligence, Vol. 00, No. 0, Month Year}
{D. Patel \MakeLowercase{\textit{et al.}}: Exploring Topic Trends in COVID-19 Research Literature using Non-Negative Matrix Factorization}

\maketitle

\begin{abstract}
In this work, we apply topic modeling using Non-Negative Matrix Factorization (NMF) on the COVID-19 Open Research Dataset (CORD-19) to uncover the underlying thematic structure and its evolution within the extensive body of COVID-19 research literature. NMF factorizes the document-term matrix into two non-negative matrices, effectively representing the topics and their distribution across the documents. This helps us see how strongly documents relate to topics and how topics relate to words. We describe the complete methodology which involves a series of rigorous pre-processing steps to standardize the available text data while preserving the context of phrases, and subsequently feature extraction using the term frequency-inverse document frequency (tf-idf), which assigns weights to words based on their frequency and rarity in the dataset. To ensure the robustness of our topic model, we conduct a stability analysis. This process assesses the stability scores of the NMF topic model for different numbers of topics, enabling us to select the optimal number of topics for our analysis. Through our analysis, we track the evolution of topics over time within the CORD-19 dataset. Our findings contribute to the understanding of the knowledge structure of the COVID-19 research landscape, providing a valuable resource for future research in this field.

\end{abstract}



\section{Introduction}

The COVID-19 pandemic has had a profound impact on global health, economies, societies, and politics \cite{covidintro, eco_conse,eco_conse1}. In response, a substantial volume of research and evidence has been published, addressing diverse aspects of the pandemic.
As highlighted in \cite{article1}, the extensive body of literature on COVID-19 necessitates specialized approaches for literature search, synthesis, and 
collaboration across a broad network of researchers. 
Various methodologies have been employed to address this challenge, including text mining techniques for managing the rapidly growing body of COVID-19 research \cite{article2}, 
the development of COVID-19 knowledge graphs \cite{article3, article5}, global analyses of early COVID-19 literature to examine research trends and geographic variations \cite{article4}, claim extraction from academic papers on COVID-19 research \cite{article9},
COVID-19 question-answering system \cite{article6}, and scientometric as well as bibliometric studies focused on COVID-19 publications \cite{article7, article8}. Understanding the evolving trends and topics within the scientific literature related to COVID-19 offers valuable insights into real-world developments and advancements in research. Topic modeling is a powerful tool for uncovering latent structures and trends within large text corpora\cite{topicmodeling_discourse}.
The COVID-19 Open Research Dataset (CORD-19) is a freely available resource of scientific papers on COVID-19 and related historical coronavirus research \cite{wang2020cord19}. CORD-19 has a rich collection of metadata and structured full text papers. The CORD-19 dataset is particularly useful for researchers interested in performing topic modeling on COVID-19 related research as it provides a vast amount of data for analysis.  By applying topic modeling techniques to the CORD-19 dataset, we can identify key themes and trends in scientific research related to COVID-19. Analyzing these trends can provide valuable insights into how the research community responds to the pandemic. Understanding the trends in the CORD-19 dataset can have far-reaching implications. Firstly, it allows us to identify active research areas and emerging topics within the scientific community \cite{journalpublication}. This knowledge can guide future research efforts and resource allocation to address critical areas related to COVID-19. Secondly, by examining the shifts in research trends over time, we can gain insights into how the understanding and priorities surrounding the pandemic have evolved \cite{topicmodeling_discourse}. This can inform policy-making, healthcare strategies, and decision-making processes in future epidemics.

Several studies have successfully applied topic modeling and temporal trend analysis to various domains  \cite{7373310}, \cite{Ghosh2017}, \cite{10.1145/2124295.2124376}, \cite{10.1007/978-981-33-6912-2_15}. For instance, Greene et al. \cite{Greene_Cross_2017} employed Non-Negative Matrix Factorization (NMF) to analyze temporal trends in the political agenda of the European Parliament. Similarly, Mishra et al. \cite{Mishra2024} conducted a temporal analysis of thematic evolution in computational economics over two decades. Several attempts have been made at topic modeling and trend analysis on scientific literature related to COVID-19. For example, Meaney et al.\cite{meaney2022} applied Non-Negative Matrix Factorization (NMF) temporal topic models to clinical text data to identify the effects of the COVID-19 pandemic on primary healthcare and community health in Toronto, Canada. Agade et al.\cite{agade2020} explored the non-medical impacts of COVID-19 using Natural Language Processing (NLP). Crane et al.\cite{crane2020} developed a new approach to identify the main topics within the COVID-19 research corpus. Urru et al.\cite{urru2022} conducted a topic trend analysis on COVID-19 literature using Structural Topic Modeling (STM).

While previous efforts have contributed significantly to the field, there is still scope for enhancement, particularly concerning the size of the datasets, the stability of topic models, and clustering effectiveness. 
For instance, Meaney et al.'s approach resulted in some highly similar topics, with identical words appearing in the highest frequency lists of multiple topics, making it sometimes difficult to label topics. Agade et al.'s NMF model could have benefited from improved clustering of topics. Crane et al.'s methodology could have been expanded to include more than 10 topics and to utilize full text articles instead of only abstracts of articles from the CORD-19 dataset. Similarly, Urru et al. could have also considered full text articles instead of solely relying on abstracts for the analysis.

To the best of our knowledge, very little to no effort has been made in the past to handle, firstly, n-grams while data cleaning, and secondly, provided a numerical stability proof of the applied model for selecting the number of topics associated with COVID-19 literature. Additionally, previous attempts are limited to the use of the most frequent words for representing topics without emphasizing on a relevance metric. This work is aimed at providing a more comprehensive view of the entire corpus while addressing the above mentioned limitations of previous works in this area. The key contributions of this article are as follows:

\begin{itemize}
\item We utilize a vast dataset of nearly 260,000 documents and apply topic modeling to the full body text of articles, enabling us to capture crucial details and nuances that previous approaches may have overlooked.

\item We employ meticulous data cleaning techniques and stability checks to ensure the reliability and qualitative interpretability of the generated topic model.

\item The words associated with each topic are carefully selected using a relevance metric, ensuring that the identified terms are not only frequent but also highly relevant to their respective topics.

\item We analyze the trends of each topic, providing a comprehensive understanding of the discourse surrounding COVID-19 and its evolution over time.
\end{itemize} 

The remainder of the article is organized as follows. In Section \ref{section2}, we discuss the process of data preparation,  specifically, the data filtering criteria and the data preprocessing steps aimed at preparing the dataset for topic modeling. 
In Section \ref{NMF}, we describe the application of NMF for identifying topics in a set of documents. We also discuss the measures and techniques employed to make sure that the NMF model is reliable and stable. Additionally, we also discuss the identification and tracking of topic trends using NMF, and finally the selection of relevant terms for topic modeling based on frequency and relevance metrics. In Section \ref{results}, we present the important topic trend analysis results while performing a comprehensive examination of the observed trends within each topic, offering valuable insights into the evolving discourse on COVID-19. Finally, in Section \ref{conclusion}, we summarize our key findings and contributions. 

\section{Data Preparation}
\label{section2}

The COVID-19 Open Research Dataset (CORD-19) \footnote{\href{https://www.kaggle.com/datasets/allen-institute-for-ai/CORD-19-research-challenge}{COVID-19 Open Research Dataset Challenge (CORD-19)} - accessed on 24th Aug 2023} is a freely available resource of over 1,000,000 scholarly articles about COVID-19, SARS-CoV-2, and related coronaviruses\cite{kaggle2020,wang2020cord19}. It was prepared by the White House and a coalition of leading research groups in response to the COVID-19 pandemic. The dataset includes over 400,000 articles with full text and is provided to the global research community to apply recent advances in NLP and other AI techniques to generate new insights in support of the ongoing fight against this infectious disease. The dataset is regularly updated to include the most recent research available, making it a valuable resource for researchers. The dataset also includes relevant information for each article as described in Table \ref{tab: feat}.

\begin{table}[h]
  \centering
  \begin{tabular}{>{\raggedright\arraybackslash}P{2.2cm} | >{\raggedright\arraybackslash}P{10cm}}
  \hline
  \multicolumn{1}{>{\centering\arraybackslash}P{1.5cm}|}{\textbf{Feature}} 
    & \multicolumn{1}{>{\centering\arraybackslash}P{6cm}}{\textbf{Description}}\\
     \hline
    cord\_uid & A unique identifier for each cord-19 paper\\
    source\_x & Sources from which the paper was received \\
    title & The corresponding paper's title\\
    doi & Document Object Identifier for the paper\\
    pmcid & A str-valued field for the paper's ID on PubMed Central\\
    pubmed\_id & An int-valued field for the paper's ID on PubMed\\
    license & A str-valued field for the most permissive license associated with the paper\\
    abstract & A str-valued field for the paper's abstract\\
    publish\_time & A str-valued field for the published date of the paper\\
    authors & A List[str]-valued field for the authors of the paper\\
    journal & A str-valued field for the paper's journal\\
    \hline
  \end{tabular}
  \caption{Relevant attributes in CORD-19 dataset }
  \label{tab: feat}
\end{table}

\subsection{Data Filtering}
\label{data_filtering}
\begin{figure*}
  \centering
  \includegraphics[width=1\textwidth]{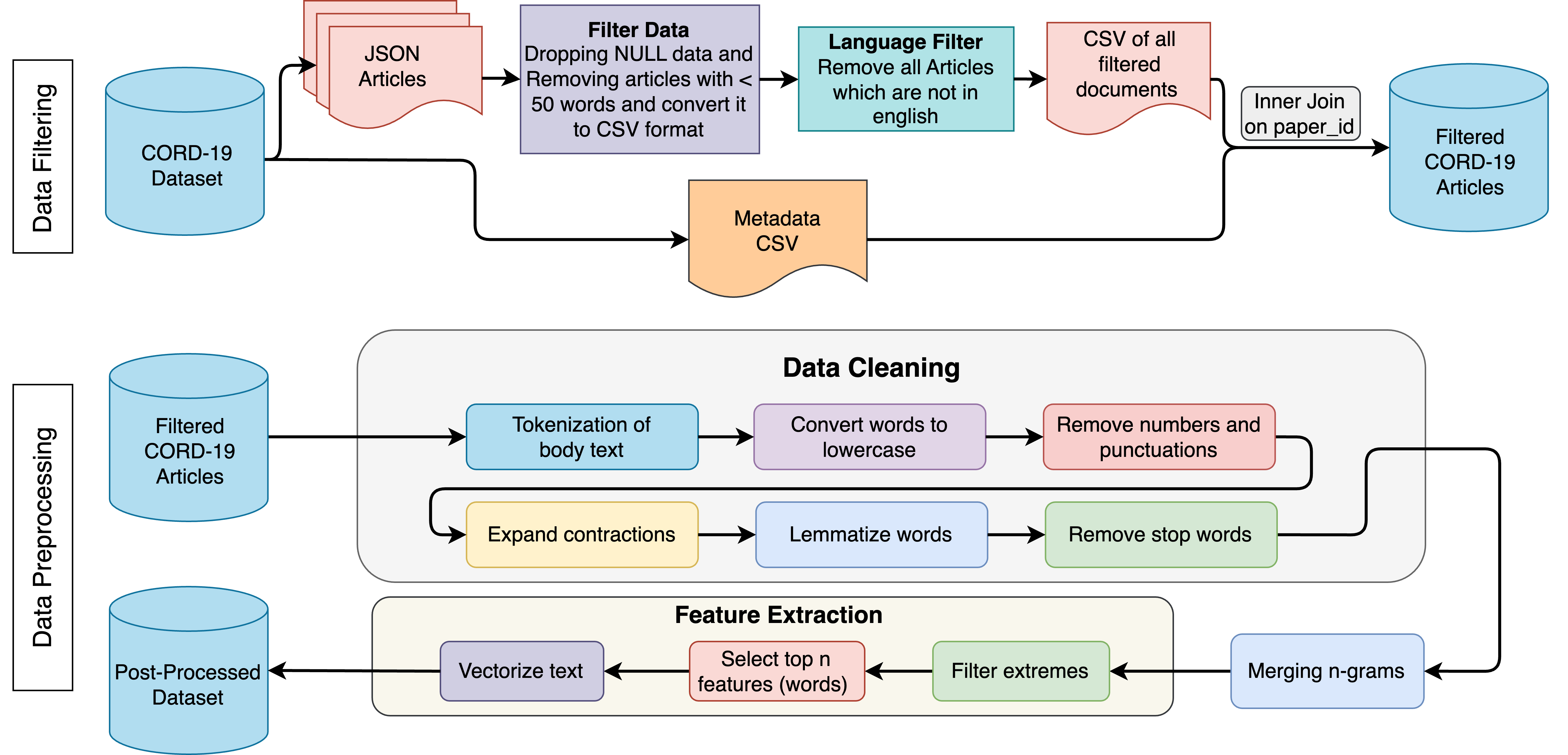}
  \caption{Methodology of data filtering and data pre-processing approach for topic modeling using NMF in the CORD-19 dataset, involving steps such as data filtering, data cleaning, n-gram handling, and feature extraction to ensure high-quality and suitable data for analysis.}
  \label{fig:preprocessing}
\end{figure*}

The data filtering process applied on the CORD-19 dataset involves several steps to ensure that the resulting corpus is of high quality and suitable for analysis, as shown in Fig. \ref{fig:preprocessing}. Firstly, articles containing less than 50 words are removed because they may not have enough context or information to provide meaningful insights when performing topic modeling. They may also be incomplete sentences or phrases that do not convey any meaning on their own or maybe the result of errors in data collection. Secondly, language detection is performed using Google's language detection library \cite{nakatani2010langdetect} to identify and keep only English articles. This ensures consistency in terms of language and allows for the effective application of NLP techniques and topic modeling algorithms for English text only. Finally, the filtered data is converted into CSV files for easier analysis. The date of publication of all articles is obtained from metadata.csv in the CORD-19 dataset. The number of papers published each month in the CORD-19 dataset varied significantly until May 2020. However, from May 2020 onward, the monthly publication rate stabilized, with roughly equal number of articles being published each month. Therefore, to ensure a consistent and representative sample for our analysis, we focus on articles published between the two year window from May 2020 to May 2022.

\begin{figure}
    \centering
    \includegraphics[width = 0.7\textwidth]{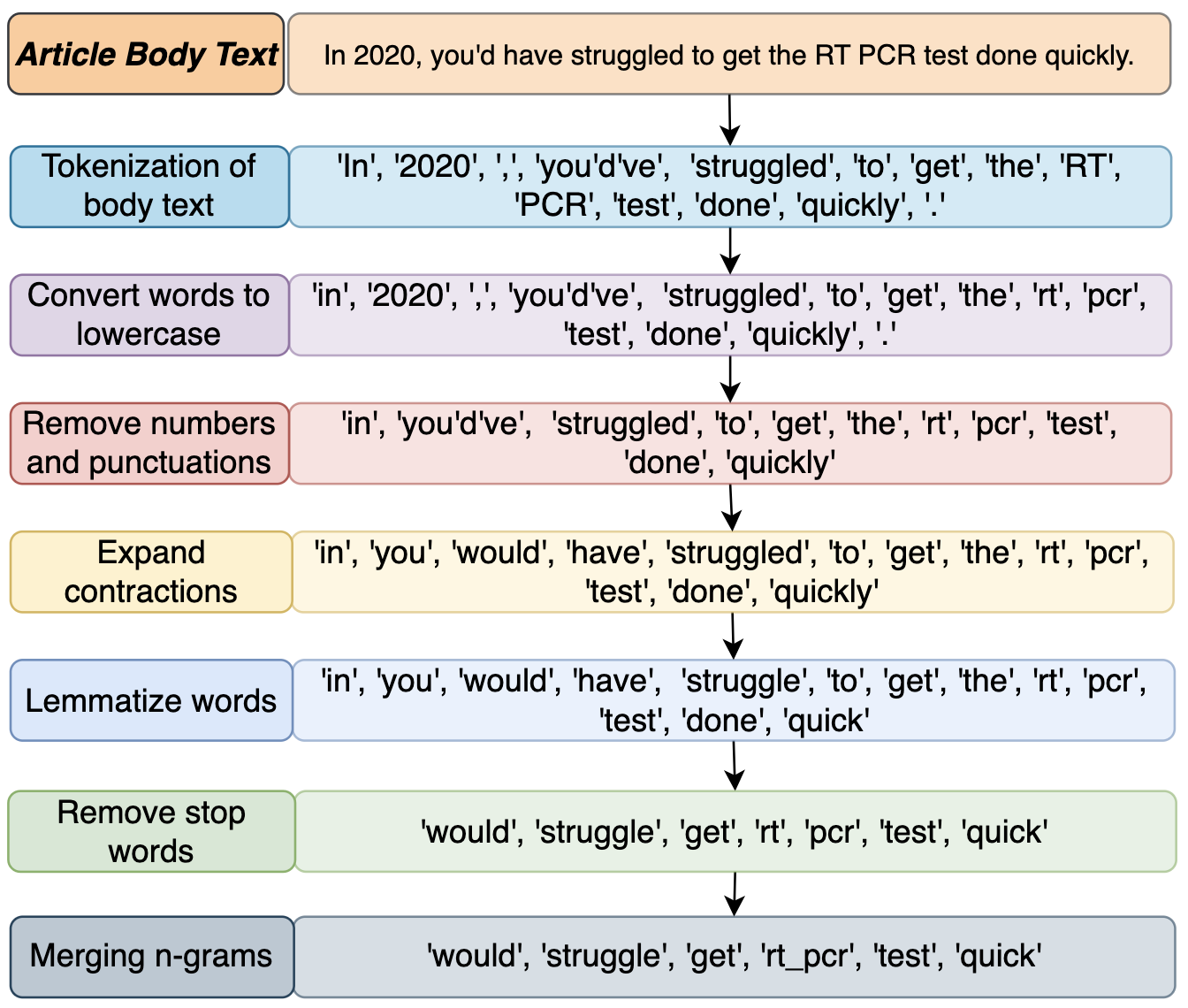}
    \caption{Example Illustrating preprocessing of article body text through the preprocessing pipeline}
    \label{fig:example-preprocessing}
\end{figure}

\subsection{Data Pre-processing}
\label{pre-processing}
The filtered CORD-19 dataset, due to its large volume, necessitates pre-processing to optimize it for topic modeling. The preprocessing steps are elaborated in the following sections and visualized in Figure \ref{fig:preprocessing}. The changes in sentences passing through this pipeline are illustrated in Figure \ref{fig:example-preprocessing}.

\subsubsection{Data Cleaning}
Data cleaning involves several preprocessing steps such as tokenization, converting words to lowercase, removing numbers and punctuation marks, contraction expansion, stop-words removal, and lemmatization to standardize and simplify text data for further analysis. Tokenization breaks down text into individual words or tokens to identify its syntactic structure and aid in further analysis. Converting words to lowercase standardizes the text and reduces its dimensionality by reducing the number of unique words. Removing numbers and punctuation marks simplifies the data by removing elements that do not carry meaning and can add noise to the data. Contraction expansion involves expanding contractions to improve the text's readability for NLP models. Stop-words removal involves removing commonly used words that do not carry much meaning to reduce the data's dimensionality and improve analysis efficiency. We developed a custom stop word list specifically tailored to domain-specific terms, including terms relevant to COVID-19 research. Lemmatization involves reducing words to their base or root form to standardize the text and understand its meaning.

\subsubsection{Merging n-grams} 
This involves identifying and merging sequences of n words that occur together in a text to provide more context to the model. This can improve the performance of the model by capturing the meaning of phrases rather than individual words. Merging n-grams allows the model to treat them as a single unit, preserving their meaning and context.

To identify relevant n-grams, we analyze the body text of articles. First, we perform frequency analysis to find the most common n-grams. Next, we manually inspect these frequent n-grams to ensure they are contextually meaningful and semantically relevant. Finally, we merge the identified n-grams into single tokens within the text corpus. For example, after performing frequency analysis, several bigrams were identified, including "receive vaccine" and "vaccine efficacy." The bigram "receive vaccine" was determined to be irrelevant and excluded, while "vaccine efficacy" was found to be meaningful and replaced with the single token "vaccine\_efficacy" throughout the text corpus. A similar process was applied to identify and merge n-grams for values of n up to 6.

By applying this approach from \textit{n = 6, 5, ..., 2}, we can identify a group of relevant bigrams, such as "vaccine hesitancy", "mental health", trigrams as "body mass index" and many more. We process n-grams in descending order (from larger n-values to smaller ones) because smaller n-grams, such as bigrams, are often components of larger n-grams, like trigrams. This method of extracting n-grams allows us to capture the nuances and contextual meanings in the text, which may not be immediately apparent through analyzing individual words. As a result, we can gain a more comprehensive understanding of the topics and themes discussed in the text. 

\subsubsection{Feature Extraction}
Feature extraction involves converting text data into numerical form for use as input for ML algorithms. This process includes creating a dictionary to remove rare and common words that are unlikely to provide useful information for the topic model. In our specific case, we apply filtering criteria to remove terms that are either too common or too rare. Specifically, we removed terms that occurred in more than 70\% of the documents or in less than 1000 (0.38\%) documents. This filtering step eliminates terms that are either too general or too specific to offer meaningful insights into the topics present in our corpus. The thresholds of 70\% and 1000 were chosen based on qualitative assessment of the corpus. Through experimentation, we observed that higher thresholds retained overly general terms, while lower thresholds excluded many informative terms. Similarly, the choice of 1000 as the minimum frequency ensured that rare but meaningful terms were retained without introducing excessive noise.

After the dictionary is created and filtering is applied, we then select the top-n words to use as features for the model. In our case, we selected the top 500,000 words based on their term frequency-inverse document frequency (tf-idf) scores \cite{ramos2003using,AIZAWA200345}. The selection of 500,000 words was determined by computational limitations, balancing the need for informative features with resource constraints. Processing more terms would significantly increase memory and computational requirements, without proportional gains in model performance. This approach gives more importance to words that are both frequent in a document and rare across the entire dataset, making them potentially more informative for our topic modeling task.

\section{Non-Negative Matrix Factorization}
\label{NMF}
Non-Negative Matrix Factorization (NMF) \cite{NIPS2000_f9d11525, lee1999learning,nmfintro} is a technique used in ML and data analysis to extract meaningful patterns and structures from high-dimensional data. In the context of topic modeling research, NMF can be used to factorize a document-term matrix (DTM) into two matrices, one representing the topics and the other representing the distribution of these topics across the documents.

Given a non-negative DTM $X \in \mathbb{R}^{m \times n}$, where $m$ is the number of documents and $n$ is the number of terms in the vocabulary, the goal of NMF is to factorize $X$ into two non-negative matrices $W \in \mathbb{R}_{\geq 0}^{m \times k}$ and $H \in \mathbb{R}_{\geq 0}^{k \times n}$, where $k$ is the number of topics.

The factorization of $X$ can be written as:

\begin{equation}
X \approx WH
\end{equation}

Each row of $W$ represents a document, and each column represents a topic. The entries in each row represent the contribution of each topic to reconstructing the corresponding document in $X$. Similarly, each column in $H$ represents a word, and each row represents a topic. The entries in each column represent the contribution of each word to reconstructing the corresponding topic. In other words, the matrix $W$ can be interpreted as a document-topic matrix, where the entries represent the strength of association between documents and topics. The matrix $H$, on the other hand, can be interpreted as a topic-word matrix, where the entries represent the strength of association between topics and words.

For parallel computation in NMF, we use the method described in \cite{kannan2016high}. This method proposes a high-performance distributed-memory parallel algorithm that computes the factorization by iteratively solving alternating non-negative least squares (NLS) subproblems for W and H. It maintains the data and factor matrices in memory (distributed across processors), uses MPI for interprocessor communication, and, in the dense case, provably minimizes communication costs. This algorithm performs well for both dense and sparse matrices. In our implementation, we utilize the source code provided by \cite{kannan2016high}, ensuring adherence to the original algorithm's design and leveraging its optimized performance for parallel computation.

\subsection{Stability Analysis}
\label{stability}

\begin{figure}[!htb]

\includegraphics[width=1.0\textwidth]{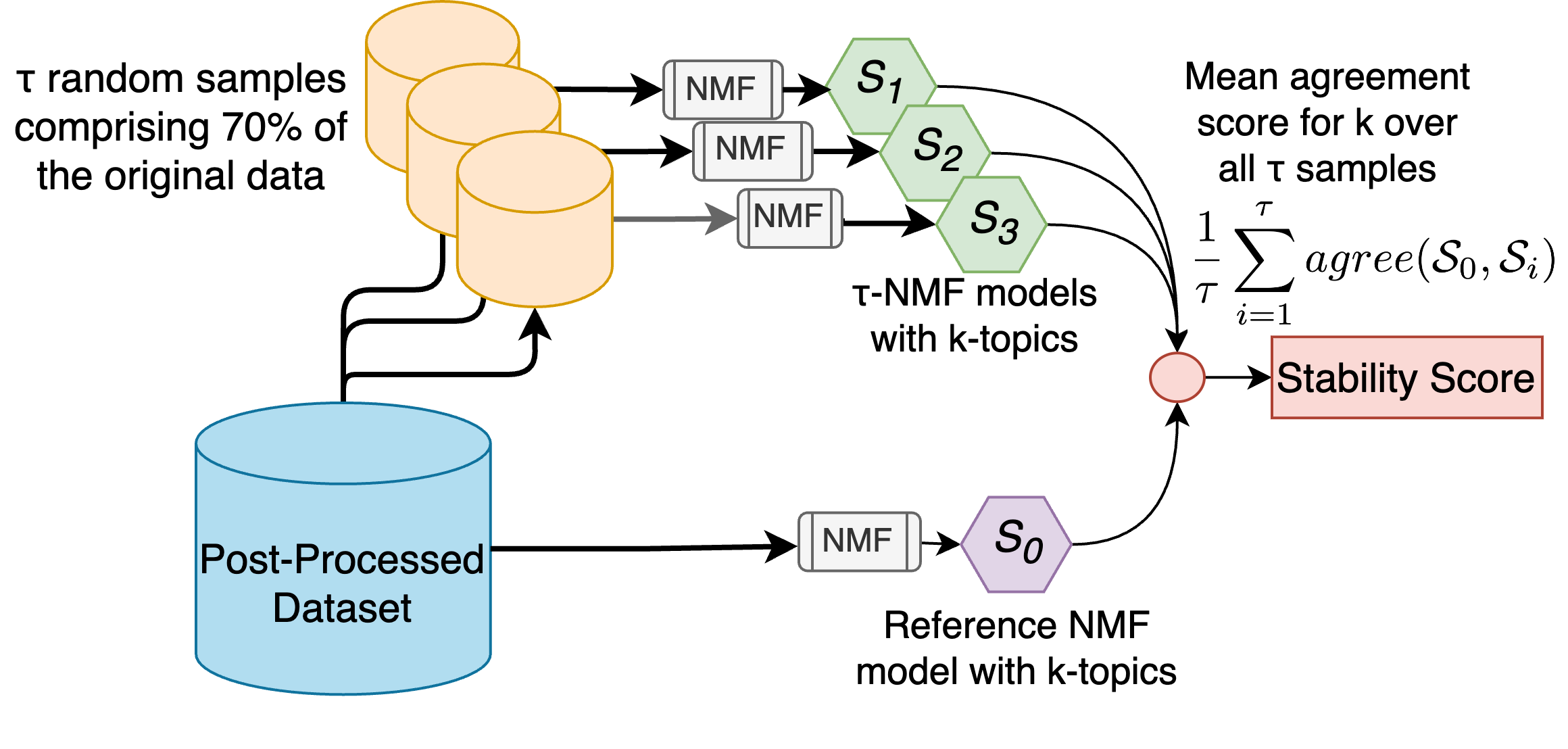}
\caption{Steps involved in assessing the stability of the topic model across multiple runs with different random initializations. }
\label{fig:Stability-illustration}
\end{figure}

One challenge in topic modeling is the inherent variability in the topics identified by the model due to its probabilistic nature and different random initializations. This variability hinders the assessment of topic model reliability and interoperability. To address this issue, we employed the stability analysis methodology proposed by Greene et al. \cite{DBLP:journals/corr/GreeneOC14}. This approach measures the similarity between multiple runs of the same topic model, each with a different random initialization, thereby providing a measure of model stability which is visualized in Figure \ref{fig:Stability-illustration}.

The output of a topic modeling algorithm is typically represented as a set of ranked lists \( S = \{R_1, \ldots, R_k\} \), where each list \( R_i \) corresponds to the \( i \)-th topic and contains the top \( t \) terms that are most indicative of that topic. For NMF, these terms are the highest valued entries in each column of the basis vectors. We use the Average Jaccard (AJ) index to measure the similarity between topic lists as follows:
\[
AJ(R_i, R_j) = \frac{1}{t} \sum_{d=1}^{t} \gamma_d(R_i, R_j),
\]
where
\[
\gamma_d(R_i, R_j) = \frac{|R_{i,d} \cap R_{j,d}|}{|R_{i,d} \cup R_{j,d}|},
\]
and \( R_{i,d} \) represents the first \( d \) terms of list \( R_i \). The AJ index is a symmetric measure that produces values between 0 and 1, with terms in a ranked list weighted on a decreasing linear scale.
To compare two distinct \( k \)-topic models, \( S_x = \{R_{x1}, \ldots, R_{xk}\} \) and \( S_y = \{R_{y1}, \ldots, R_{yk}\} \), we construct a similarity matrix \( M \) of size \( k \times k \). Each entry \( M_{ij} \) reflects the AJ index between the \( i \)-th topic of the first model and the \( j \)-th topic of the second model. Optimal pairing between the two sets of topics is determined by solving the minimal weight bipartite matching problem using the Hungarian method, resulting in an agreement score:
\[
\text{agree}(S_x, S_y) = \frac{1}{k} \sum_{i=1}^{k} AJ(R_{xi}, \pi(R_{xi})),
\]
where \( \pi(R_{xi}) \) is the list in \( S_y \) that best matches \( R_{xi} \) according to the permutation \( \pi \). The agreement score ranges from 0 to 1, with 1 indicating perfect agreement between two identical \( k \)-topic models. For \( \tau \) different samples, the overall stability is the average of the agreement scores calculated for each pair of samples:
\[
\text{stability} = \frac{1}{\tau} \sum_{i=1}^{\tau} \text{agree}(S_i, S_0).
\]

\begin{figure}[!htb]
\includegraphics[width=1.0\textwidth]{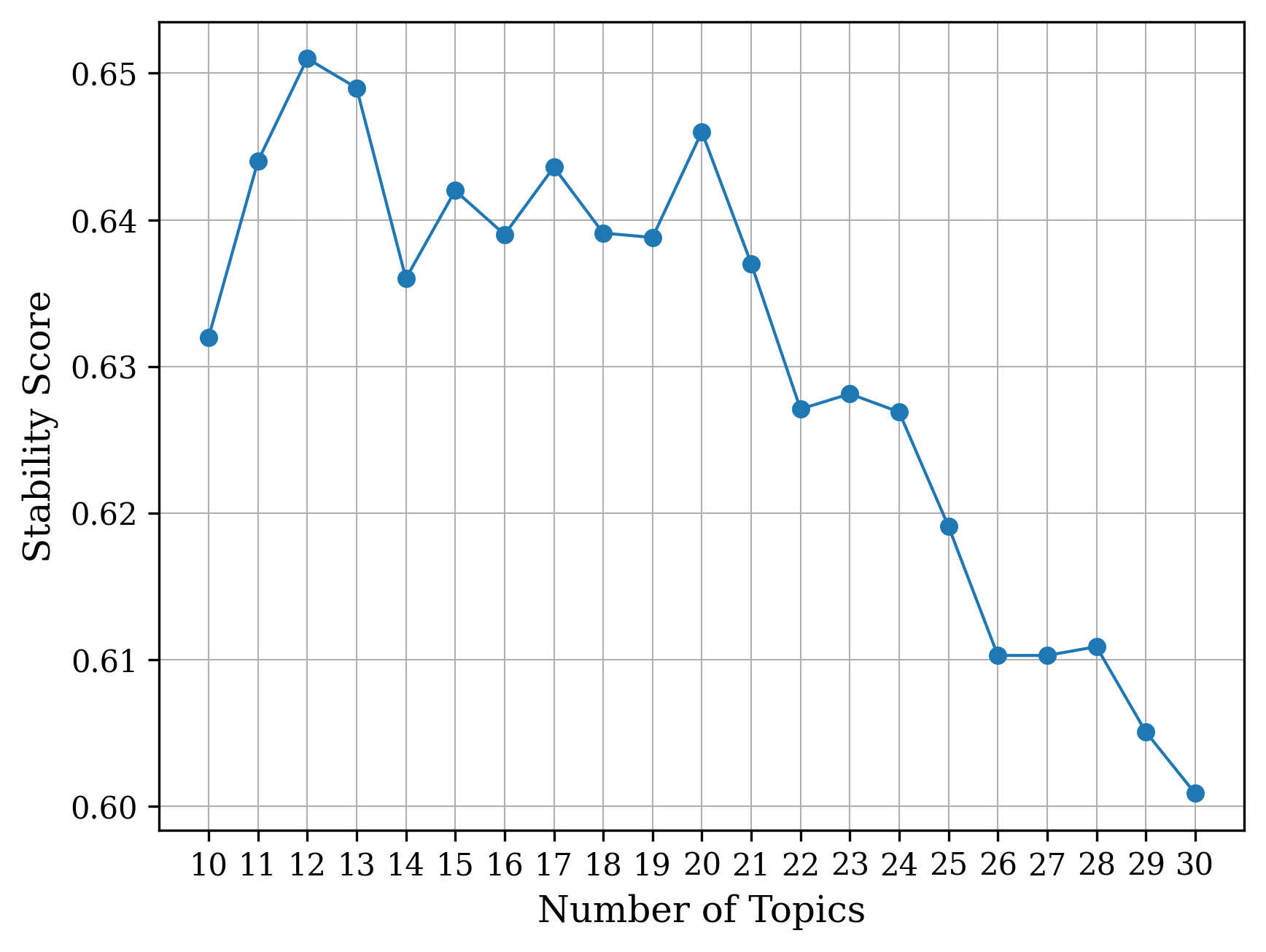}
\caption{Stability scores of NMF topic model for different numbers of topics }
\label{fig:stability plot}
\end{figure}

Our experimental results demonstrated consistently high stability scores for our topic model, even when the number of topics exceeded 25. This indicates that our model consistently identifies a stable set of topics, even with a large number of topics. A possible explanation for this result is that our dataset encompasses a wide range of distinct and well-separated topics. In such cases, allowing the model to identify more topics enhances its ability to capture the underlying structure of the data, leading to the derivation of more stable and interpretable topics. Figure \ref{fig:stability plot} shows the stability scores of our NMF topic model for different numbers of topics.

These findings provide good evidence for the robustness and reliability of our topic model. The consistent identification of a stable set of topics across multiple runs increases the confidence in the interpretability and generalizability of the topics identified by our model. Based on the stability analysis, we selected a topic model with 20 topics for further analysis. Additionally, we conducted a qualitative assessment of the top words associated with each topic after running the model. This assessment revealed that the words strongly align with meaningful and coherent themes, further validating the quality of the topics produced by our model. 

While increasing the number of topics beyond 20 may uncover additional distinct topics, however their impact on the overall data structure is relatively minor and does not significantly enhance the interpretability or generalizability of our model. For instance, when we increased the number of topics to 40, our topic model identified additional topics, such as a separate topic for sleep-related problems, a separate topic for stroke, and a separate topic for stem cells. Although these topics are distinct and well-separated, our qualitative assessment of their associated words indicates that their contributions to the overall understanding of the dataset are relatively small. In other words, the presence of these additional topics does not significantly improve the interpretability or generalizability of our model.
We believe, based on our qualitative assessment, the choice of 20 topics strikes a balance between model complexity and interpretability.

\subsection{Identifying Topic Trends using NMF}
\label{Topic-trends}
To investigate topic trends over time, we partitioned the article collection obtained after filtering and pre-processing the CORD-19 dataset by month. Let $n_k$ denote the total number of articles published in month $k$, and $I_k$ represent the set of indices of articles belonging to month $k$. Given a matrix $W$ of dimensions $n \times t$, where $n$ is the total number of articles and $t$ is the number of topics, we calculate the average topic distribution for each month as follows:
\begin{equation}
T_k = \frac{1}{n_k} \sum_{i \in I_k} \frac{W_i}{\sum_{l=1}^{t}W_{il}}
\end{equation}
here, $W_i$ denotes the $i^{th}$ row of matrix $W$, and $W_{il}$ represents the entry at the $l^{th}$ column of the $i^{th}$ row of $W$. We normalize each row by dividing each element by the sum of all elements in that row. These normalized values are then summed for all articles published in a given month and divided by the total number of articles to obtain the average topic distribution, $T_k$, for that month.\\
By analyzing the average topic distributions for each month, we can identify trends in the evolution of topics over time and gain insights into how different topics change and interact with one another.
\begin{figure}[!htb]
\includegraphics[width=1.0\textwidth]{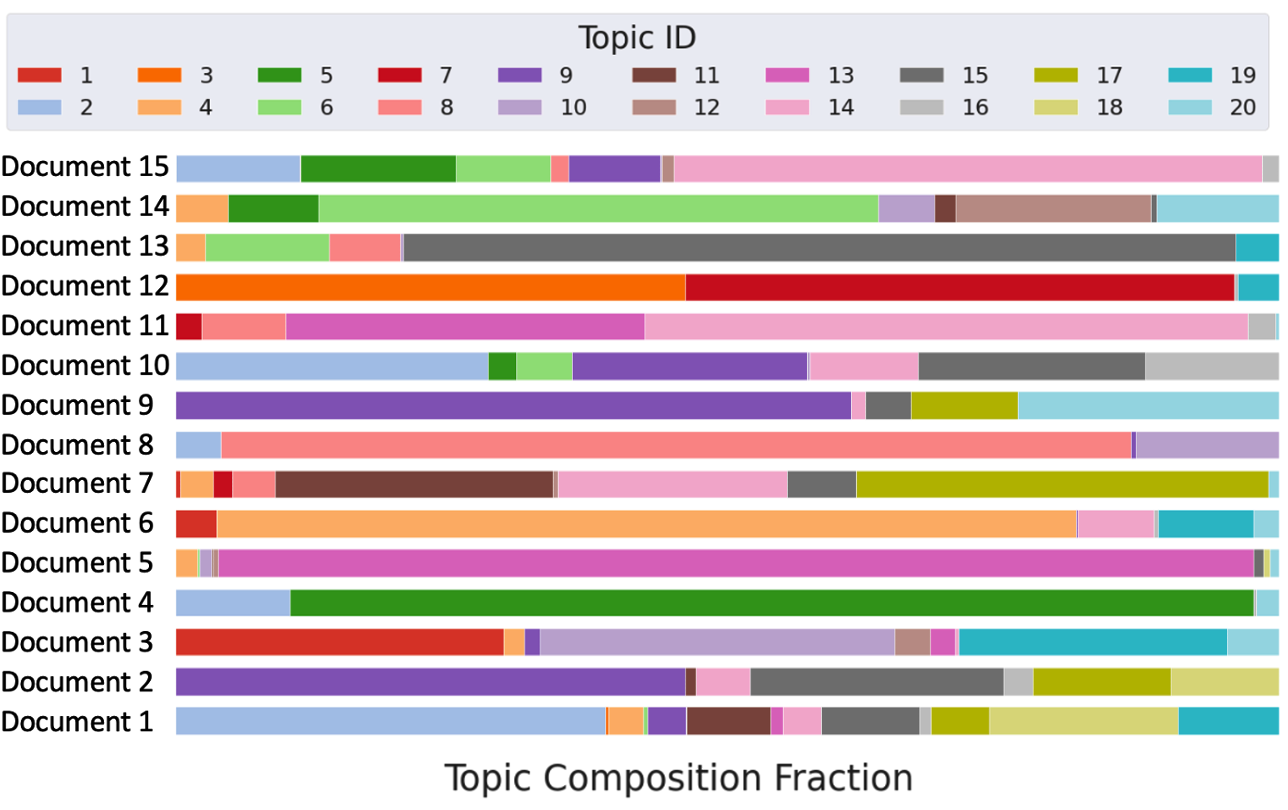}
\caption{Illustration of the distribution of topic fractions in a random sample of 25 documents}
\label{fig:Distribution}
\end{figure}
Each document is characterized by a mixture of various topics. This mixture is graphically represented in Figure \ref{fig:Distribution}. The contribution of each topic to a particular document is quantified by the entries present in the corresponding row of a matrix $W$. By repeating this process for all months, we can analyze how topics change over time and uncover trends in topic evolution. Representing each document as a mixture of topics, rather than assigning a single topic to any document, allows us to capture the complexity and richness of the content. Documents often contain multiple themes and ideas, and representing them as mixtures enables us to more accurately model their content. This approach also allows us to identify relationships between topics and analyze how they interact with one another over time. By employing this approach, we can enhance our understanding of topic trends and gain valuable insights into how different topics evolve and interact with one another over time.

\subsection{Finding Relevant Terms for Topic Modeling on CORD-19 Dataset using Relevance}
\label{relevance-form}
An important challenge in topic modeling is identifying relevant terms for each topic. To address this challenge, we employ the relevance formula proposed by Sievert and Shirley \cite{sievert-shirley-2014-ldavis}. The relevance of a term $w$ to a topic $t$ is denoted as $R(w \mid t)$ and is defined as:
\begin{equation}
R(w \mid t) = \lambda p(w \mid t) + (1-\lambda) \frac{p(w \mid t)}{p(w)}
\label{eq: relevance_formula}
\end{equation}
\\
The relevance formula, $R(w \mid t)$, ranks terms within topics based on their relevance scores. $\lambda$ is a weight parameter that balances the probability of term $w$ under topic $t$ and its lift. Lift measures how often the term appears in that topic compared to its overall frequency in the corpus. Adjusting $\lambda$ controls the importance given to the probability of a term under a specific topic relative to its lift. When $\lambda = 1$, terms are ranked solely based on their probability under a specific topic, and when $\lambda = 0$, terms are ranked solely based on their lift. For intermediate values of $\lambda$ between 0 and 1, both probability and lift are considered when ranking terms.

High-frequency words that appear in multiple topics may not provide much information about a specific topic because they are common across the entire corpus. By considering both the probability of a term under a specific topic and its lift, the relevance formula can help identify terms that are more informative and valuable for understanding each topic.
By employing the relevance formula, we can enhance the interpretability and effectiveness of NMF topic modeling on the CORD-19 dataset, enabling the identification of relevant terms that contribute to the characterization and understanding of each topic.
In our implementation, we use $\lambda = 0.5$ to balance the contributions of term probability and lift, giving equal weight to both factors. This ensures that the relevance formula considers both the specificity of a term to a topic and its overall prominence in the corpus, enabling a more nuanced selection of relevant terms.

\section{Results}
\label{results}
In our analysis, we used a corpus of 261,342 unique documents. These documents have been preprocessed as discussed in previous sections and transformed the text data into a document-term matrix (DTM) using the term frequency-inverse document frequency (tf-idf) weighting scheme as discussed in \ref{pre-processing}. Next, we applied NMF to factorize the DTM matrix into two matrices, representing the topic-term and document-topic relationships. In our analysis, we specified the number of topics to be 20, resulting in a topic-term matrix $H$ with dimensions $20 \times 500,000$ and a document-topic matrix $W$ with dimensions $261,342 \times 20$. 
After performing NMF, we applied the relevance formula (\ref{eq: relevance_formula}) with $\lambda$ set to 0.5 to rank the terms for each topic. This formula balanced the probability of a term under a specific topic and its lift, providing a measure of the term's informativeness for that topic. Table \ref{Table:Top Words} presents the top relevant words for each topic, allowing us to gain insights into the key themes represented by the topics.

\begin{table*}[h]
\centering
\caption{Top words for each topic in the CORD-19 dataset identified using the relevance formula introduced by Sievert and Shirley \cite{sievert-shirley-2014-ldavis} for $\lambda = 0.5$.}
\label{Table:Top Words}
\renewcommand{\arraystretch}{1.2}
\footnotesize
\begin{tabular}{p{0.6cm} | p{3.0cm} | p{9.0cm}}
\hline
\textbf{No.} & \textbf{Topic Label} & \textbf{Top Relevant Words} \\
\hline
1  & Hospitalization and Comorbidities & icu, admission, diabetes, aki, mechanical\_ventilation, comorbidities, hospitalized, admitted, anticoagulation, oxygen \\
\hline
2  & Economic Policy and Crisis & food, economic, sector, business, policy, economy, firm, political, price, investment  \\
\hline
3  & Vaccines and Vaccination & vaccine, vaccination, vaccine\_hesitancy, dose, bnt, pfizer, dos, mrna\_vaccine, immunization, biontech \\
\hline
4  & Education and Online Learning & student, learning, teaching, online\_learning, instructor, classroom, lecture, medical\_student, online\_teaching, semester \\
\hline
5  & Immune Response and Inflammation & il, cytokine, inflammatory, macrophage, ifn, tnf, activation, mouse, monocyte, inflammation \\
\hline
6  & Mental Health and Well-being & anxiety, depression, mental\_health, psychological, stress, sleep, depressive, coping, loneliness, mental \\
\hline
7  & Antibody Response and Testing & antibody, igg, igm, iga, elisa, serum, rbd, titer, neutralization, antibody\_response \\
\hline
8  & Children & child, parent, school, adolescent, pediatric, parenting, childhood, caregiver, mother, preschool \\
\hline
9  & Machine Learning & image, dataset, algorithm, layer, classification, accuracy, neural\_network, classifier, deep\_learning, architecture \\
\hline
10 & Telemedicine and Healthcare & telemedicine, telehealth, visit, provider, team, staff, consultation, nurse, resident, virtual \\
\hline
11 & Virus Variants and Mutations & mutation, variant, genome\_sequence, lineage, clade, omicron, gisaid, phylogenetic, delta \\
\hline
12 & PCR Testing and RNA Detection & rt\_pcr, specimen, swab, saliva, apcr, rna, ct\_value, lamp, assay \\
\hline
13 & Pregnancy and Maternal Health & woman, pregnancy, maternal, birth, trimester, preterm, mother, postpartum, neonatal, gestational \\
\hline
14 & Airborne Transmission and Mask Usage & mask, aerosol, air\_particle, droplet, pm, airborne, respirator, temperature, filtration, face\_mask \\
\hline
15 & Lung Imaging and Pneumonia & ct, chest, consolidation, lung, scan, opacity, ggo, pneumonia, lesion, lobe \\
\hline
16 & Epidemic Control and Transmission & epidemic, lockdown, reproduction\_number, sir, susceptible, mobility, seir, equation, confirmed\_case, cumulative \\
\hline
17 & Cancer & cancer, surgery, chemotherapy, tumor, surgical, oncology, breast\_cancer, radiotherapy, colorectal, lung\_cancer \\
\hline
18 & Drug Design and Antiviral Compounds & compound, ligand, docking, residue, kcal\_mol, molecule, nsp, atom, binding, hydrogen\_bond \\
\hline
19 & ACE2 Receptor and Virus Binding & ace, ang, arb, ang\_ii, rbd, ace\_expression, angiotensin, tmprss, ace\_protein, raas \\
\hline
20 & Healthcare Workers & hcws, hcw, ppe, nurse, ipc, occupational, worker, frontline, seroprevalence, healthcare\_worker \\
\hline
\end{tabular}
\end{table*}

Upon the completion of term ranking for each topic, we proceed to label the topics. This is a crucial step in topic modeling as it provides a human-interpretable summary of each topic. The labeling has been performed manually by examining the top relevant words for each topic, as presented in Table \ref{Table:Top Words}. To ensure the validity and reliability of these labels, we adopted a collaborative approach involving the authors of this study and volunteers. Each individual independently suggested labels for each topic based on their interpretation of the top relevant words. Following the independent labeling, a discussion was held to reconcile any differences and reach a consensus on the most suitable label for each topic. This method has been adopted to reduce bias and enhance the robustness of the labels.

\subsection{Topic Trend Analysis}
\label{topic_trend_analysis}
\begin{figure*}[!h]
\centering
\includegraphics[width=1\textwidth]{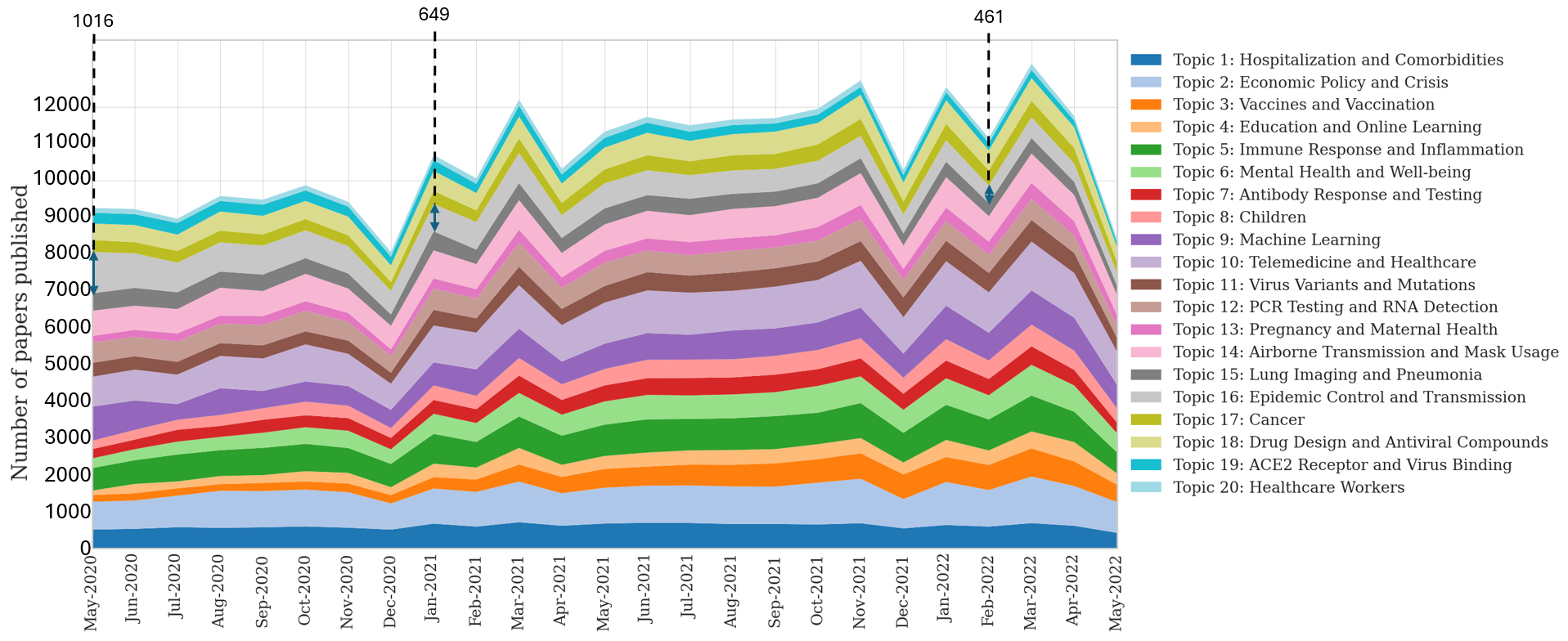}
\caption{The figure shows temporal trends in the number of research articles across 20 distinct COVID-19-related topics from May 2020 to May 2022. It illustrates both the cumulative total and the relative contributions of each topic, highlighting shifts in research focus over time. Notable trends include the emergence and decline of interest in specific areas over time such as vaccination, drug design, epidemic control, mental health and telemedicine. 
The dotted line illustrates the changing prominence of \textit{Topic 16: Epidemic Control and Transmission}, as indicated by variations in relative width over time.}
\label{fig:trend streamgraph}
\end{figure*}

Our analysis of topic trends during the COVID-19 pandemic from May 2020 to May 2022 revealed several interesting patterns. As shown in Figure \ref{fig:trend streamgraph}, topics such as vaccination, education and online learning, mental health and well-being, children, telemedicine, and healthcare exhibited an increasing trend over time. In contrast, topics such as lung imaging and pneumonia, epidemic control and transmission, PCR testing and RNA detection, airborne transmission and mask usage, and ACE2 receptor and virus binding displayed a decreasing trend. Other topics, such as hospitalization and comorbidities, immune response and inflammation, cancer, drug design and antiviral compounds, and healthcare workers remained relatively stable during this period.

\begin{figure*}[!h]
  \centering
  \subfloat[]{\includegraphics[width=0.44\textwidth]{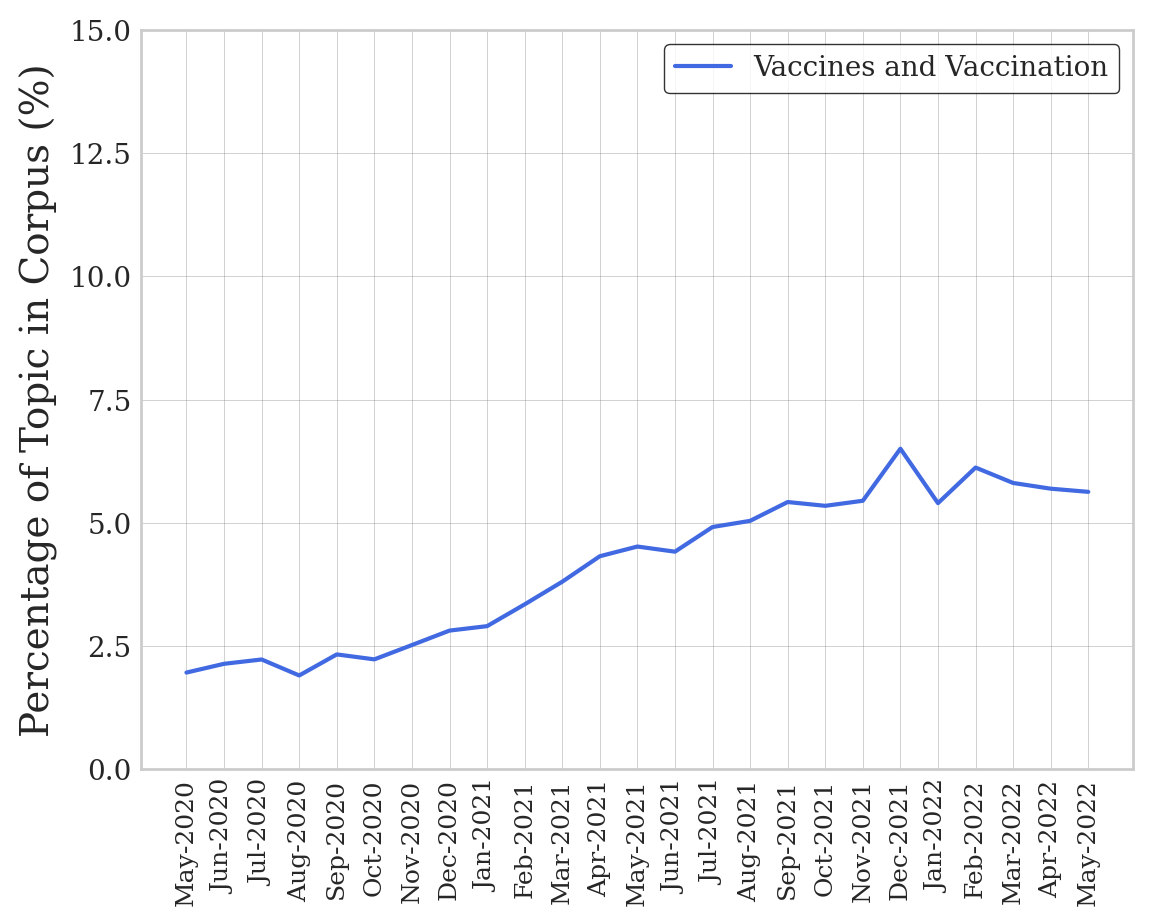}\label{fig:sub1}}
  \hspace{0.05\textwidth}
  \subfloat[]{\includegraphics[width=0.44\textwidth]{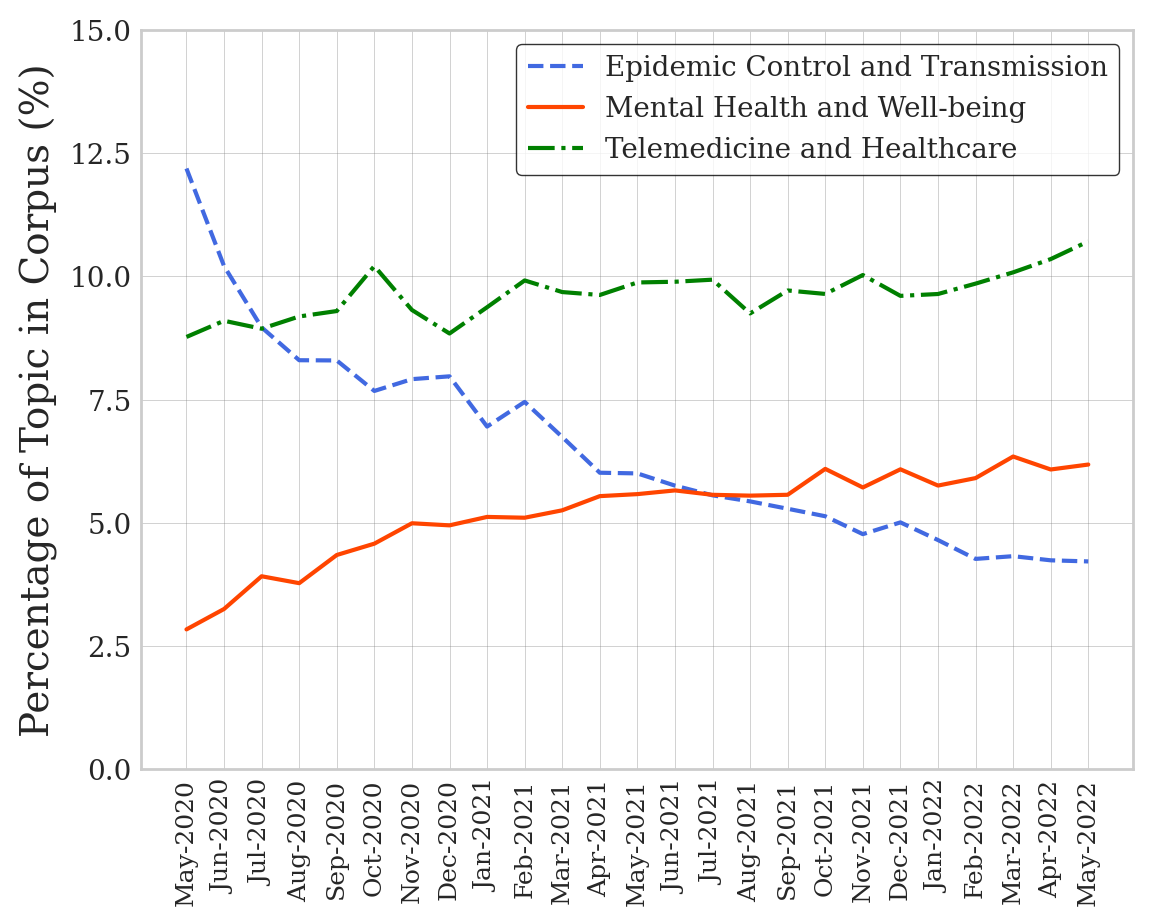}\label{fig:sub2}}\\
  \subfloat[]{\includegraphics[width=0.44\textwidth]{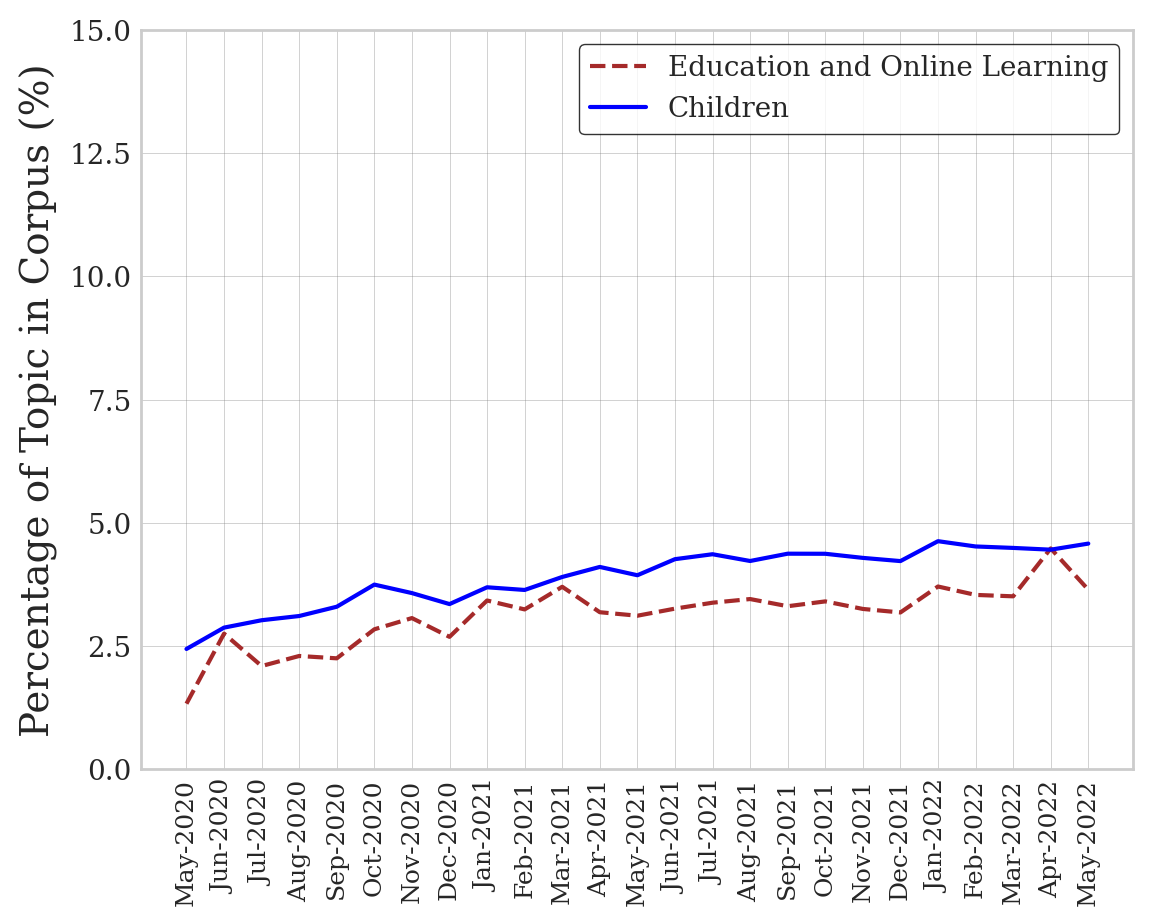}\label{fig:sub3}}
  \hspace{0.05\textwidth}
  \subfloat[]{\includegraphics[width=0.44\textwidth]{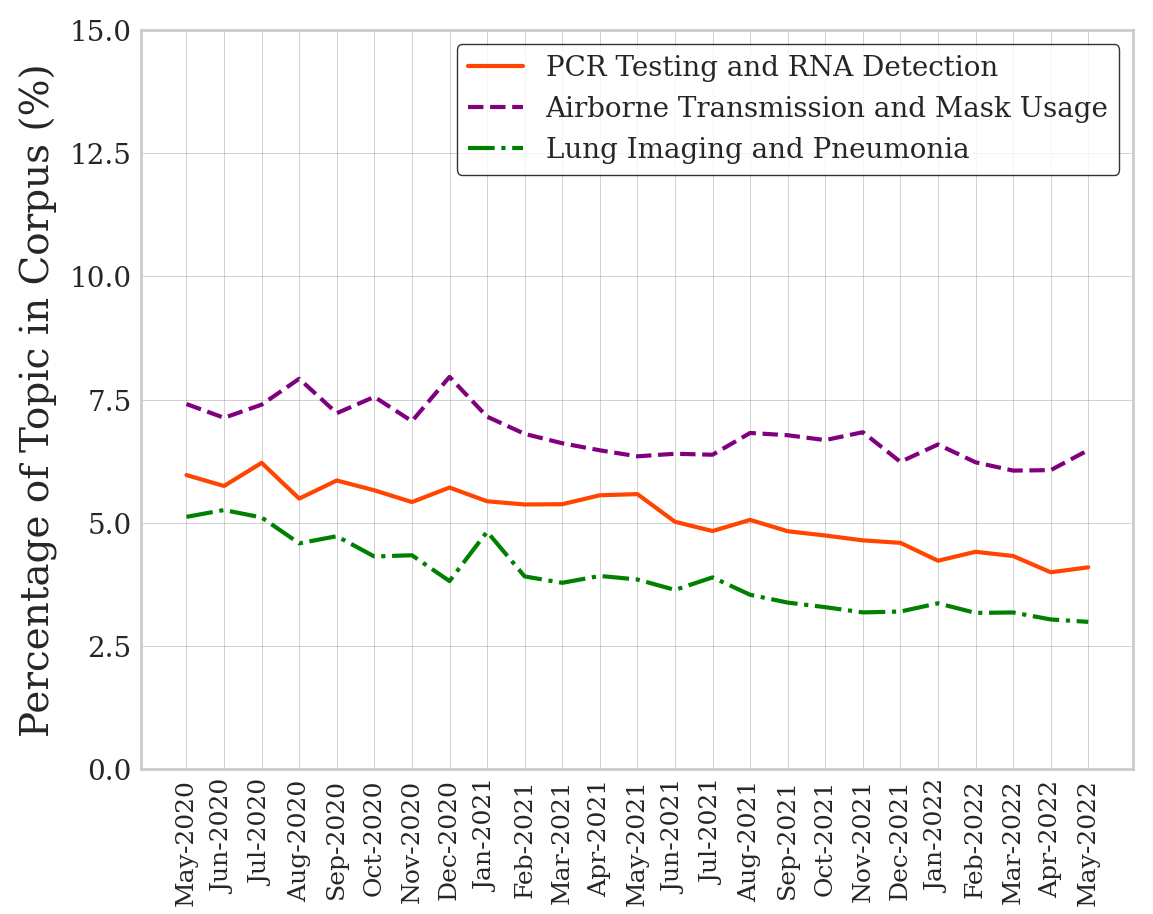}\label{fig:sub4}}
  \caption{Temporal Evolution of Key Topics in COVID-19 Research literature: The graphs represent the temporal trends in research publications for select prominent topics during the COVID-19 pandemic.}
  \label{fig:plot_grid}
\end{figure*}

As depicted in Figure \ref{fig:trend streamgraph} and Figure \ref{fig:plot_grid}(a) there has been a significant increase in research publications related to vaccines and vaccination during the COVID-19 pandemic. This surge in research output reflects the heightened importance and focus on vaccine development and deployment, as well as updates regarding vaccine efficacy and safety \cite{mallapaty2021covid}. The escalating trend, observed from September 2020 onwards, might be attributed to the pressing need for effective vaccines to mitigate the spread of the disease, reduce fatalities, and alleviate the burden on healthcare systems worldwide. The pharmaceutical and research sectors have been compelled to expedite vaccine development due to the urgency and gravity of the situation \cite{mallapaty2021covid}. Additionally, studies have been conducted to analyze perceptions of vaccine safety and promise in various populations \cite{vaccineanalysis}. Furthermore, research has also focused on understanding the host immune response to COVID-19 \cite{immuneresponse}.

There are interconnected trends observed during the COVID-19 pandemic, encompassing Mental Health and Well-being, Telemedicine and Healthcare, and Epidemic Control and Transmission, as shown in Figure \ref{fig:plot_grid}(b). The increasing trends in mental health and well-being, and Telemedicine\cite{demeke2021trends} and healthcare could be attributed to the profound impact of the pandemic on physical and mental health \cite{who2022mental}, necessitating the adoption of novel approaches for healthcare delivery \cite{monaghesh2020role}. The COVID-19 pandemic has accelerated key technology trends, including telehealth, which can help reduce the spread of the coronavirus while helping businesses stay open \cite{weforum2020tech}. This acceleration is reflected in the increasing trend in Telemedicine and Healthcare. Moreover, the pandemic has exposed persistent inequalities by income, age, race, sex, and geographic location. Despite recent global health gains, across the world, people continue to face complex, interconnected threats to their health and well-being rooted in social, economic, political, and environmental determinants of health \cite{who2021impact}. This has led to an increased focus on Mental Health and Well-being as individuals seek information and resources to cope with the pandemic's emotional toll \cite{covidmentalhealth,covidmentalhealth2}. Consequently, we hypothesize that the trend for Epidemic Control and Transmission has shown a gradual decline as people have acclimated to preventive measures such as social distancing and mask usage. Moreover, we speculate that the diminishing trend in Epidemic Control and Transmission is also influenced by the widespread vaccination efforts, shifting the focus from containment to management of the impact of the virus. This shift has probably prompted a greater emphasis particularly on Mental Health and Well-being, and also on Telemedicine and Healthcare as revealed in Figure \ref{fig:plot_grid}(b).

Figure \ref{fig:plot_grid}(c) underscores the significant shift towards online learning and its implications for child development during the COVID-19 pandemic. We observe an escalating trend in discussions related to child development and online learning, which can be attributed to several factors. Firstly, we hypothesize that widespread school closures, a necessary measure to curb the spread of the virus, have necessitated the adoption of online learning platforms. This sudden transition has raised concerns among parents, educators, and policymakers regarding its impact on children's cognitive, social, and emotional development \cite{Li2021}. Secondly, the pandemic may have expedited the integration of digital technology in education. This acceleration has prompted discussions on the effectiveness of online learning \cite{10.1371/journal.pone.0273016}, digital literacy skills, and the digital divide among students. Additionally, we speculate that parental involvement in children's education has increased as a result of the transition to online learning. With children studying from home, parents may have become more involved in their children's education, further influencing discussions around this topic.

The evolving trends in discussions surrounding RT-PCR Testing \cite{Yuan2006}, Airborne Transmission \cite{10.1093/occmed/kqaa080} and Mask Usage \cite{LI2020141560}, Lung Imaging, and Pneumonia \cite{Rousan2020} throughout the COVID-19 pandemic, as depicted in Figure \ref{fig:plot_grid}(d), can be attributed to various factors. The prominent factor is the increasing adoption of vaccination. Firstly, the widespread availability of vaccines has reduced the need for frequent RT-PCR Testing, particularly among vaccinated individuals. As vaccine coverage expands, the necessity for testing diminishes, leading to a decline in discussions related to this topic. Secondly, vaccines provide protection against severe illness and hospitalization, consequently reducing the need for discussions on lung imaging and pneumonia. As more individuals receive the vaccine, the risk of severe illness diminishes, thereby decreasing the emphasis on related topics. Lastly, the increasing trend in vaccine adoption has probably resulted in a reduced focus on airborne transmission and mask usage, as vaccines effectively reduce the likelihood of transmission. Consequently, discussions surrounding these preventive measures have decreased.\\
This comprehensive analysis of topic trends during the COVID-19 pandemic highlights the increasing research focus on vaccines and vaccination, the interrelated dynamics between Mental Health and Well-being, Telemedicine and Healthcare, and Epidemic Control and Transmission, the shift towards online learning and its impact on child development, as well as the changing discourse around testing, transmission, and preventive measures due to the widespread vaccination efforts. These findings provide valuable insights into the evolving landscape of research and public discourse in response to the unprecedented challenges posed by the pandemic.

\section{Conclusion}
\label{conclusion}

In conclusion, our analysis of the CORD-19 dataset using NMF topic modeling has provided valuable insights into the evolution of topics related to the COVID-19 pandemic. By characterizing and monitoring these topic trends over time, we have been able to identify changing priorities and concerns among researchers. Our findings demonstrate the potential of NMF topic modeling as a scalable methodology for analyzing large collections of unstructured text data. The stability of our results, while descriptive and hypothesis-generating, provides valuable information that can complement other sources of data and assist stakeholders in understanding the diverse impacts of the COVID-19 pandemic.

Furthermore, our approach can be applied to other pandemics or public health crises to analyze trends in scientific publications and inform decision-making. By leveraging the unique and expressive information contained in scientific publications, we can enhance our understanding of public health challenges and inform planning and resource allocation.

Our analysis also revealed that the trends extracted from the CORD-19 dataset align with real-world events and developments. For instance, the increasing trend in discussions related to vaccines and vaccination corresponds with the global urgency for effective vaccines to combat COVID-19. Similarly, the escalating trend in discussions related to online learning reflects the widespread school closures necessitated by the pandemic. 
Overall, our analysis provides a powerful tool for monitoring and understanding the evolution of topics related to the COVID-19 pandemic and other public health challenges. It also imparts valuable information about post-pandemic planning and resource allocation, thereby contributing to more effective response to future public health crises. Future research could expand the dataset to include sources like news articles and social media posts. This would provide more context and insights, especially for understanding public perception and misinformation during pandemics. Integrating diverse data types may offer a more comprehensive view of societal responses to health crises.

\section*{Credit Authorship Contribution Statement}
\textbf{Divya Patel}: Conceptualization, Methodology, Investigation, Validation, Data curation, Software, Data Analysis, Writing - original draft,  Writing - review \& editing. \textbf{Vansh Parikh}: Conceptualization, Methodology, Investigation, Validation, Data curation, Software, Writing - original draft,  Writing - review \& editing. \textbf{Om Patel}: Conceptualization, Investigation, Validation, Data curation, Software. \textbf{Agam Shah}: Conceptualization, Methodology, Supervision, Investigation, Data Analysis, Writing - review \& editing. \textbf{Bhaskar Chaudhury}: Conceptualization, Methodology, Investigation, Validation, Data Analysis and Visualization, Supervision, Writing - original draft, Writing - review \& editing.

\section*{Declaration of Competing Interest}
The authors declare that they have no known competing financial interests or personal relationships that could have appeared to influence the work reported in this paper.

\section*{Funding}
This research did not receive any specific grant from funding agencies in the public, commercial, or not-for-profit sectors.

\nocite{*}
\bibliographystyle{abbrv}
\bibliography{references}

\begin{thebibliography}{10}

\bibitem{agade2020}
A.~Agade and S.~Balpande.
\newblock Exploring the non-medical impacts of covid-19 using natural language processing.
\newblock {\em Preprints.org 2020}, page 2020110056, 2020.

\bibitem{AIZAWA200345}
A.~Aizawa.
\newblock An information-theoretic perspective of tf–idf measures.
\newblock {\em Information Processing \& Management}, 39(1):45--65, 2003.

\bibitem{journalpublication}
S.~X. Bian and E.~Lin.
\newblock Competing with a pandemic: Trends in research design in a time of covid-19.
\newblock {\em PLOS ONE}, 15(9):1--14, 09 2020.

\bibitem{10.5555/944919.944937}
D.~M. Blei, A.~Y. Ng, and M.~I. Jordan.
\newblock Latent dirichlet allocation.
\newblock {\em J. Mach. Learn. Res.}, 3(null):993–1022, mar 2003.

\bibitem{10.1093/occmed/kqaa080}
J.~Borak.
\newblock {Airborne Transmission of COVID-19}.
\newblock {\em Occupational Medicine}, 70(5):297--299, 06 2020.

\bibitem{7373310}
Y.~Chen, H.~Zhang, J.~Wu, X.~Wang, R.~Liu, and M.~Lin.
\newblock Modeling emerging, evolving and fading topics using dynamic soft orthogonal nmf with sparse representation.
\newblock In {\em 2015 IEEE International Conference on Data Mining}, pages 61--70, 2015.

\bibitem{article7}
G.~Colavizza, R.~Costas, V.~Traag, N.~J. van Eck, T.~Van~Leeuwen, and L.~Waltman.
\newblock A scientometric overview of cord-19.
\newblock {\em PLOS ONE}, 16:e0244839, 01 2021.

\bibitem{crane2020}
A.~Crane, B.~Freidrich, W.~Fehlman, I.~Frolow, and D.~W. Engels.
\newblock A novel methodology to identify the primary topics contained within the covid-19 research corpus.
\newblock {\em SMU Data Science Review}, 3(2):Article 1, 2020.

\bibitem{demeke2021trends}
H.~B. Demeke, S.~Merali, S.~Marks, L.~Z. Pao, L.~Romero, P.~Sandhu, H.~Clark, A.~Clara, K.~B. McDow, E.~Tindall, et~al.
\newblock Trends in use of telehealth among health centers during the covid-19 pandemic — united states, june 26–november 6, 2020.
\newblock {\em Morbidity and Mortality Weekly Report}, 70(7):240--244, 2021.

\bibitem{9737322}
P.~Deng, T.~Li, H.~Wang, D.~Wang, S.-J. Horng, and R.~Liu.
\newblock Graph regularized sparse non-negative matrix factorization for clustering.
\newblock {\em IEEE Transactions on Computational Social Systems}, 10(3):910--921, 2023.

\bibitem{article3}
D.~Domingo-Fernández, S.~Baksi, B.~Schultz, Y.~Gadiya, R.~Karki, T.~Raschka, C.~Ebeling, M.~Hofmann-Apitius, and A.~Kodamullil.
\newblock Covid-19 knowledge graph: a computable, multi-modal, cause-and-effect knowledge model of covid-19 pathophysiology.
\newblock {\em Bioinformatics (Oxford, England)}, 37, 09 2020.

\bibitem{article8}
H.~Elhawary, A.~Salimi, N.~Diab, and L.~Smith.
\newblock Bibliometric analysis of early covid-19 research: The top 50 cited papers.
\newblock {\em Infectious Diseases: Research and Treatment}, 13, 10 2020.

\bibitem{covidmentalhealth}
A.~S. Feroz, N.~A. Ali, N.~A. Ali, R.~Feroz, S.~N. Meghani, and S.~Saleem.
\newblock Impact of the covid-19 pandemic on mental health and well-being of communities: an exploratory qualitative study protocol.
\newblock {\em BMJ Open}, 10(12), 2020.

\bibitem{weforum2020tech}
W.~E. Forum.
\newblock 10 tech trends getting us through the covid-19 pandemic.
\newblock \url{https://www.weforum.org/agenda/2020/04/10-technology-trends-coronavirus-covid19-pandemic-robotics-telehealth}, 2020.

\bibitem{Ghosh2017}
S.~Ghosh, P.~Chakraborty, E.~O. Nsoesie, E.~Cohn, S.~R. Mekaru, J.~S. Brownstein, and N.~Ramakrishnan.
\newblock Temporal topic modeling to assess associations between news trends and infectious disease outbreaks.
\newblock {\em Scientific Reports}, 7(1):40841, January 19 2017.

\bibitem{covidmentalhealth2}
N.~S. Gray, C.~O'Connor, J.~Knowles, J.~Pink, N.~J. Simkiss, S.~D. Williams, and R.~J. Snowden.
\newblock The influence of the covid-19 pandemic on mental well-being and psychological distress: Impact upon a single country.
\newblock {\em Frontiers in Psychiatry}, 11, 2020.

\bibitem{Greene_Cross_2017}
D.~Greene and J.~P. Cross.
\newblock Exploring the political agenda of the european parliament using a dynamic topic modeling approach.
\newblock {\em Political Analysis}, 25(1):77–94, 2017.

\bibitem{DBLP:journals/corr/GreeneOC14}
D.~Greene, D.~O'Callaghan, and P.~Cunningham.
\newblock How many topics? stability analysis for topic models.
\newblock {\em CoRR}, abs/1404.4606, 2014.

\bibitem{eco_conse1}
T.~Ibn-Mohammed, K.~Mustapha, J.~Godsell, Z.~Adamu, K.~Babatunde, D.~Akintade, A.~Acquaye, H.~Fujii, M.~Ndiaye, F.~Yamoah, and S.~Koh.
\newblock A critical analysis of the impacts of covid-19 on the global economy and ecosystems and opportunities for circular economy strategies.
\newblock {\em Resources, conservation, and recycling}, 164:105169, 01 2021.

\bibitem{topicmodeling_discourse}
W.~Jo.
\newblock Possibility of discourse analysis using topic modeling.
\newblock {\em Journal of Asian Sociology}, 48(3):321--342, 2019.

\bibitem{kaggle2020}
Kaggle.
\newblock Covid-19 open research dataset challenge (cord-19).
\newblock \url{https://www.kaggle.com/allen-institute-for-ai/CORD-19-research-challenge}, 2020.

\bibitem{kannan2016high}
R.~Kannan, G.~Ballard, and H.~Park.
\newblock A high-performance parallel algorithm for nonnegative matrix factorization.
\newblock In {\em Proceedings of the 21st ACM SIGPLAN Symposium on Principles and Practice of Parallel Programming}, page~7, 2016.

\bibitem{NIPS2000_f9d11525}
D.~Lee and H.~S. Seung.
\newblock Algorithms for non-negative matrix factorization.
\newblock In T.~Leen, T.~Dietterich, and V.~Tresp, editors, {\em Advances in Neural Information Processing Systems}, volume~13. MIT Press, 2000.

\bibitem{lee1999learning}
D.~D. Lee and H.~S. Seung.
\newblock Learning the parts of objects by non-negative matrix factorization.
\newblock {\em Nature}, 401(6755):788--791, 1999.

\bibitem{Li2021}
C.~Li and F.~Lalani.
\newblock How covid-19 has changed the way we educate children.
\newblock \url{https://www.weforum.org/agenda/2021/02/digital-learning-covid-19-changed-way-we-educate-children/}, 2021.

\bibitem{LI2020141560}
Y.~Li, R.~Zhang, J.~Zhao, and M.~J. Molina.
\newblock Understanding transmission and intervention for the covid-19 pandemic in the united states.
\newblock {\em Science of The Total Environment}, 748:141560, 2020.

\bibitem{covidintro}
Y.-C. Liu, R.-L. Kuo, and S.-R. Shih.
\newblock Covid-19: The first documented coronavirus pandemic in history.
\newblock {\em Biomedical Journal}, 43(4):328--333, 2020.

\bibitem{mallapaty2021covid}
S.~Mallapaty, E.~Callaway, M.~Kozlov, H.~Ledford, J.~Pickrell, and R.~Van~Noorden.
\newblock How covid vaccines shaped 2021 in eight powerful charts.
\newblock {\em Nature}, 598, 2021.

\bibitem{meaney2022}
C.~Meaney, M.~Escobar, R.~Moineddin, T.~A. Stukel, S.~Kalia, B.~Aliarzadeh, T.~Chen, B.~O'Neill, and M.~Greiver.
\newblock Non-negative matrix factorization temporal topic models and clinical text data identify covid-19 pandemic effects on primary healthcare and community health in toronto, canada.
\newblock {\em Journal of Biomedical Informatics}, 128:104034, 2022.

\bibitem{Mishra2024}
M.~Mishra, S.~K. Vishwakarma, L.~Malviya, and S.~Anjana.
\newblock Temporal analysis of computational economics: a topic modeling approach.
\newblock {\em International Journal of Data Science and Analytics}, July 2024.

\bibitem{monaghesh2020role}
E.~Monaghesh and A.~Hajizadeh.
\newblock The role of telehealth during covid-19 outbreak: a systematic review based on current evidence.
\newblock {\em BMC Public Health}, 20(1):1193, 2020.

\bibitem{article1}
P.~Nogueira, M.~J. Forjaz, C.~Rodriguez-Blazquez, A.~Diaz-Franco, B.~Unim, L.~Palmieri, L.~Carcaillon-Bentata, T.~Makovski, and R.~Feteira-Santos.
\newblock Research methodologies to assess the impact of covid-19.
\newblock {\em European Journal of Public Health}, 31, 10 2021.

\bibitem{telemedicine}
OECD.
\newblock {\em The COVID-19 Pandemic and the Future of Telemedicine}.
\newblock 2023.

\bibitem{who2021impact}
W.~H. Organization.
\newblock The impact of covid-19 on global health goals.
\newblock \url{https://www.who.int/news-room/spotlight/the-impact-of-covid-19-on-global-health-goals}, 2021.

\bibitem{who2022mental}
W.~H. Organization.
\newblock Covid-19 and mental health: a review of the existing literature.
\newblock \url{https://www.who.int/publications/i/item/WHO-2019-nCoV-Sci_Brief-Mental_health-2022.1}, 2022.

\bibitem{vaccineanalysis}
V.~C. Osuji, E.~M. Galante, D.~Mischoulon, J.~E. Slaven, and G.~Maupome.
\newblock Covid-19 vaccine: A 2021 analysis of perceptions on vaccine safety and promise in a u.s. sample.
\newblock {\em PLOS ONE}, 17(5):1--19, 05 2022.

\bibitem{eco_conse}
A.~Pak, O.~Adegboye, A.~Adekunle, K.~Rahman, E.~Mcbryde, and D.~Eisen.
\newblock Economic consequences of the covid-19 outbreak: the need for epidemic preparedness.
\newblock {\em Frontiers in Public Health}, 8:1--4, 05 2020.

\bibitem{article5}
S.~Pestryakova, D.~Vollmers, A.~Sherif, S.~Heindorf, M.~Saleem, D.~Moussallem, and A.-C. Ngonga~Ngomo.
\newblock Covidpubgraph: A fair knowledge graph of covid-19 publications.
\newblock {\em Scientific Data}, 9:389, 07 2022.

\bibitem{ramos2003using}
J.~Ramos et~al.
\newblock Using tf-idf to determine word relevance in document queries.
\newblock In {\em Proceedings of the first instructional conference on machine learning}, volume 242, pages 29--48. Citeseer, 2003.

\bibitem{article6}
S.~Raza, B.~Schwartz, and L.~Rosella.
\newblock Coquad: a covid-19 question answering dataset system, facilitating research, benchmarking, and practice.
\newblock {\em BMC Bioinformatics}, 23, 06 2022.

\bibitem{Rousan2020}
L.~A. Rousan, E.~Elobeid, M.~Karrar, and Y.~Khader.
\newblock Chest x-ray findings and temporal lung changes in patients with covid-19 pneumonia.
\newblock {\em BMC Pulmonary Medicine}, 20(1):245, 2020.

\bibitem{10.1145/2124295.2124376}
A.~Saha and V.~Sindhwani.
\newblock Learning evolving and emerging topics in social media: a dynamic nmf approach with temporal regularization.
\newblock In {\em Proceedings of the Fifth ACM International Conference on Web Search and Data Mining}, WSDM '12, page 693–702, New York, NY, USA, 2012. Association for Computing Machinery.

\bibitem{article9}
D.~Shah, K.~Shah, M.~Jagani, A.~Shah, and B.~Chaudhury.
\newblock Concord: enhancing covid-19 research with weak-supervision based numerical claim extraction.
\newblock {\em Journal of Intelligent Information Systems}, pages 1--23, 09 2024.

\bibitem{nakatani2010langdetect}
N.~Shuyo.
\newblock Language detection library for java, 2010.

\bibitem{sievert-shirley-2014-ldavis}
C.~Sievert and K.~Shirley.
\newblock {LDA}vis: A method for visualizing and interpreting topics.
\newblock In {\em Proceedings of the Workshop on Interactive Language Learning, Visualization, and Interfaces}, pages 63--70, Baltimore, Maryland, USA, June 2014. Association for Computational Linguistics.

\bibitem{10.1007/978-981-33-6912-2_15}
S.~Sivanandham, A.~Sathish~Kumar, R.~Pradeep, and R.~Sridhar.
\newblock Analysing research trends using topic modelling and trend prediction.
\newblock In V.~S. Reddy, V.~K. Prasad, J.~Wang, and K.~T.~V. Reddy, editors, {\em Soft Computing and Signal Processing}, pages 157--166, Singapore, 2021. Springer Singapore.

\bibitem{article4}
B.~Tran, G.~Ha, L.~Hoang, G.~Vu, M.~Hoang, H.~Le, C.~Latkin, and C.~Ho.
\newblock Studies of novel coronavirus disease 19 (covid-19) pandemic: A global analysis of literature.
\newblock {\em International Journal of Environmental Research and Public Health}, 17:4095, 06 2020.

\bibitem{urru2022}
S.~Urru, V.~Sciannameo, C.~Lanera, S.~Salaris, D.~Gregori, and P.~Berchialla.
\newblock A topic trend analysis on covid-19 literature.
\newblock {\em DIGITAL HEALTH}, 8, 2022.

\bibitem{9745326}
D.~Wang, T.~Li, P.~Deng, J.~Liu, W.~Huang, and F.~Zhang.
\newblock A generalized deep learning algorithm based on nmf for multi-view clustering.
\newblock {\em IEEE Transactions on Big Data}, 9(1):328--340, 2023.

\bibitem{10.1145/3584862}
D.~Wang, T.~Li, P.~Deng, F.~Zhang, W.~Huang, P.~Zhang, and J.~Liu.
\newblock A generalized deep learning clustering algorithm;based on non-negative matrix factorization.
\newblock {\em ACM Trans. Knowl. Discov. Data}, 17(7), may 2023.

\bibitem{WANG2023101884}
D.~Wang, T.~Li, W.~Huang, Z.~Luo, P.~Deng, P.~Zhang, and M.~Ma.
\newblock A multi-view clustering algorithm based on deep semi-nmf.
\newblock {\em Information Fusion}, 99:101884, 2023.

\bibitem{article2}
L.~Wang and K.~Lo.
\newblock Text mining approaches for dealing with the rapidly expanding literature on covid-19.
\newblock {\em Briefings in Bioinformatics}, 22, 12 2020.

\bibitem{wang2020cord19}
L.~L. Wang, K.~Lo, Y.~Chandrasekhar, R.~Reas, J.~Yang, D.~Eide, et~al.
\newblock Cord-19: The covid-19 open research dataset.
\newblock {\em ArXiv200407761 Cs}, 2020.

\bibitem{10071530}
Y.~Wang, W.~Shi, Y.~Sun, and C.-H. Yeh.
\newblock A novel framework to forecast covid-19 incidence based on google trends search data.
\newblock {\em IEEE Transactions on Computational Social Systems}, 11(1):1352--1361, 2024.

\bibitem{nmfintro}
Y.-X. Wang and Y.-J. Zhang.
\newblock Nonnegative matrix factorization: A comprehensive review.
\newblock {\em IEEE Transactions on Knowledge and Data Engineering}, 25(6):1336--1353, 2013.

\bibitem{10.3389/fcomm.2021.651997}
P.~Wicke and M.~M. Bolognesi.
\newblock Covid-19 discourse on twitter: How the topics, sentiments, subjectivity, and figurative frames changed over time.
\newblock {\em Frontiers in Communication}, 6, 2021.

\bibitem{immuneresponse}
Y.~Xia, R.-q. Yao, P.-y. Zhao, Z.-b. Tao, L.-y. Zheng, H.-t. Zhou, Y.-m. Yao, and X.-m. Song.
\newblock Publication trends of research on covid-19 and host immune response: A bibliometric analysis.
\newblock {\em Frontiers in Public Health}, 10, 2022.

\bibitem{10.3389/fmed.2021.743988}
N.~Yin, S.~Dellicour, V.~Daubie, N.~Franco, M.~Wautier, C.~Faes, D.~Van~Cauteren, L.~Nymark, N.~Hens, M.~Gilbert, M.~Hallin, and O.~Vandenberg.
\newblock Leveraging of sars-cov-2 pcr cycle thresholds values to forecast covid-19 trends.
\newblock {\em Frontiers in Medicine}, 8, 2021.

\bibitem{Yuan2006}
J.~S. Yuan, A.~Reed, F.~Chen, and C.~N. Stewart.
\newblock Statistical analysis of real-time pcr data.
\newblock {\em BMC Bioinformatics}, 7(1):85, 2006.

\bibitem{10.1371/journal.pone.0273016}
J.~Zhang, Y.~Ding, X.~Yang, J.~Zhong, X.~Qiu, Z.~Zou, Y.~Xu, X.~Jin, X.~Wu, J.~Huang, and Y.~Zheng.
\newblock Covid-19’s impacts on the scope, effectiveness, and interaction characteristics of online learning: A social network analysis.
\newblock {\em PLOS ONE}, 17(8):1--21, 08 2022.

\bibitem{zhu2022}
L.~Zhu and S.~W. Cunningham.
\newblock Unveiling the knowledge structure of technological forecasting and social change (1969-2020) through an nmf-based hierarchical topic model.
\newblock {\em Technological Forecasting and Social Change}, 174:121277, 2022.

\end{thebibliography}

\end{document}